\documentclass{article}

\usepackage{PRIMEarxiv}

\usepackage[utf8]{inputenc} 
\usepackage[T1]{fontenc}    
\usepackage{hyperref}       
\usepackage{url}            
\usepackage{booktabs}       
\usepackage{amsfonts}       
\usepackage{nicefrac}       
\usepackage{microtype}      
\usepackage{lipsum}
\usepackage{fancyhdr}       
\usepackage{graphicx}       
\usepackage{natbib}
\graphicspath{{media/}}     

\title{Metric Effects based on Fluctuations in values of k in Nearest Neighbor Regressor}

\author{
  Abhishek Gupta \\
  Department of EXTC \\
  University of Mumbai \\
  Mumbai, 400032, India \\
  \texttt{abhishekgupta20001@gmail.com} \\
   \And
  Raunak Joshi \\
  Department of IT \\
  University of Mumbai \\
  Mumbai, 400032, India \\
  \texttt{raunakjoshi.m@gmail.com} \\
  \And
  Nandan Kanvinde \\
  Department of MCA \\
  TIMSCDR \\
  Mumbai, 400101, India \\
  \texttt{kanvindenandan81@gmail.com} \\
  \And
  \AND
  Pinky Gerela \\
  Assistant Professor - Dept. of MCA \\
  TIMSCDR \\
  Mumbai, 400101, India \\
  \texttt{dr.pinkyg5@gmail.com} \\
  \And
  Ronald Melwin Laban \\
   Assistant Professor - Dept. of EXTC \\
   St. John College of Engineering and Management \\
   Palghar, 401404, India \\
   \texttt{ronaldlaban@gmail.com} \\
}

\begin{document}
\maketitle

\begin{abstract}
Regression branch of Machine Learning purely focuses on prediction of continuous values. The supervised learning branch has many regression based methods with parametric and non-parametric learning models. In this paper we aim to target a very subtle point related to distance based regression model. The distance based model used is K-Nearest Neighbors Regressor which is a supervised non-parametric method. The point that we want to prove is the effect of k parameter of the model and its fluctuations affecting the metrics. The metrics that we use are Root Mean Squared Error and R-Squared Goodness of Fit with their visual representation of values with respect to k values.
\end{abstract}


\section{Introduction}
Machine Learning branch has made a great advancement in terms of improvement and adaptability with respect to different form of data. The Machine Learning has Classification \citep{10.2307/2344237} and Regression \citep{Maulud_Abdulazeez_2020} as major learning methodologies. These later have subdivisions of Supervised \citep{10.5555/1566770.1566773} and Unsupervised \citep{Lngkvist2014ARO} learning. The Regression is used for the prediction of continuous values and Classification for discrete value prediction. The learning methods are divided in parametric and non-parametric \citep{7321316} type models. Considering the parametric models the types of algorithms are Linear Regression \citep{10.1371/journal.pone.0262918}, Logistic Regression \citep{cramer2002origins}, Discriminant Analysis \citep{gupta2022discriminant} whereas in non-parametric models, the types are K Nearest Neighbors \citep{9065747}, Support Vector Machines \citep{708428}, Decision Trees \citep{10.1023/A:1022643204877}, Bagging Ensemble Methods \citep{kanvinde2022binary}, Boosting Ensemble \citep{gupta2021succinct} Methods. Regression is something we are aiming for this paper. The regression has many varied models, viz. Linear Regression, Lasso Regression \citep{10.2307/2346178}, Ridge Regression \citep{10.2307/1271436} which are linear models. Similarly there are regression based models with Support Vector Machines, Bagging Ensemble and Distance based algorithms. This paper is not a compendious comparison but is the subtle observation of the distance based learning method known as K Nearest Neighbor Regressor. The algorithm is supervised and non-parametric. The point that we are trying to prove in this paper is related to effect of fluctuations in the k value over regression based metrics. The metrics used are going to be Root Mean Squared Error and Goodness of Fit measure known as R-Squared Score.

\section{Methodology}
K-Nearest Neighbors abbreviated as KNN is a Supervised Learning algorithm. It is non-parametric approach to density estimation \citep{10.2307/4144429} that makes few assumptions about the form of the distribution. It requires labelled input data which is given to the model for training and later test on validation data to compare with the expected output. Kernel Width is an important parameter and is denoted using $h$. When density of data is high, the data is smoothened to the extent of deleting the invaluable insight from the data. If there is reduction in $h$, the noise in the data increases hence giving a vague prediction and landing the prediction in the area of less predictive section. Thus the optimal choice for $h$ may be depend on location inside the data space where need of KNN arises.

\begin{equation}
    P(X) = \frac{K}{NV}
    \label{eq:1}
\end{equation}

The Equation \ref{eq:1} gives the probability distribution where $K$ are number of parameters and $V$ is volume. The equation gives the general result for local density estimation, instead of fixing $V$, consideration for $K$ is done. The KNN algorithm uses \textit{feature similarity} to predict the values on newly observed data points. This explains that the new point is assigned a value based on how closely it resembles the points in the training set. The judgement of the algorithm is done using distance based formulas. These distance based formulas are required to distinguish the data points efficiently one from each other. 

\subsection{Euclidean Distance}
Euclidean distance \citep{Liberti2014EuclideanDG} is calculated as the square root of the sum of the squared differences between a new observation and an existing observation. The formula for this can be given as

\begin{equation}
    d(x,y) = \sqrt{\sum_i^n (x_i-y_i)^2}
    \label{eq:2}
\end{equation}

The Equation \ref{eq:2} has $x$ and $y$ as its 2 points for calculation where summation is taken over $n$ range of numbers. A square is taken for avoiding the negative values.

\subsection{Manhattan Distance}
Manhattan Distance \citep{Ranjitkar2016ComparisonOA} is the distance between real vectors using the sum of their absolute difference. The formula can be given as

\begin{equation}
    d(x,y) = \sum_i^n |x_i-y_i|
    \label{eq:3}
\end{equation}

The Equation \ref{eq:3} is a representation of the Manhattan Distance where the 2 values taken into consideration are $x$ and $y$ and an absolute values are taken to avoid the negative values.

\subsection{Hamming Distance}
Hamming Distance \citep{NIPS2012_59b90e10} is used for categorical variables. The formula representation is exactly as same as Manhattan Distance with set of imposed rules. If the value $x$ and the value $y$ are the similar, the distance is equal to 0 else it will be 1.

\subsection{Regressor}
The K Nearest Neighbors Regressor operates on the principles of K Nearest Neighbors at every instance where the significant arbitrary values declared by the user as a parameter is known as $k$. Nearby points have more influence on the regression that points that faraway in weight nearest neighbors. 

\subsection{Dataset}
The problem is subtle observation of the K Nearest Regressor so the horizon for selecting datasets was very vast for us. We decided to perform the method with multiple standardized datasets. The datasets we used are regression based datasets, which are Boston Housing Prices, QSAR fish toxicity LC50 and CO2 Emission by Vehicles. These are widely used regression based datasets and should suffice to serve the purpose of this paper.

\section{Results}
This section of the paper gives the outcomes that we have achieved after implementation. The results can be measured in a broad sense using regression based metrics. The metrics are explained below along with the performed results.

\subsection{Root Mean Squared Error}
As the words make up the metrics, the meaning of it could be understood word by word. The first consideration should be given to Error. The error is also known as Residual. Residuals are a measure of how far from the regression line data points are separated. These are nothing but prediction error which is literally subtracting the predicted value from actual value. This error is squared later to avoid the negative values. This can later be used to give the Sum of Squared Errors which is a summation of all the squared errors. The formula for it can be represented as

\begin{equation}
    SSE = \sum_i^n (y_i-\hat{y_i})^2
    \label{eq:4}
\end{equation}

The Equation \ref{eq:4} considers the true label and predicted label. The summation of all such values are taken over the range of $n$ values. Now mean is calculated which gives Mean Squared Error \citep{sammut2011encyclopedia} which is metric for Regression based models. The formula for Mean Squared Error is given by

\begin{equation}
    MSE = \frac{1}{n}\sum_i^n (y_i-\hat{y_i})^2
\end{equation}

Further the Root of the Mean Squared Error is calculated by formula

\begin{equation}
    RMSE = \sqrt{\frac{\sum_i^n (y_i-\hat{y_i})^2}{n}}
\end{equation}

In RMSE \citep{gmd-7-1247-2014}, before average the residuals are squared. This is an indication that RMSE is useful when large residuals are present and they do affect the performance of the model. 

\begin{figure}[htbp]
    \centering
    \includegraphics[scale=0.4]{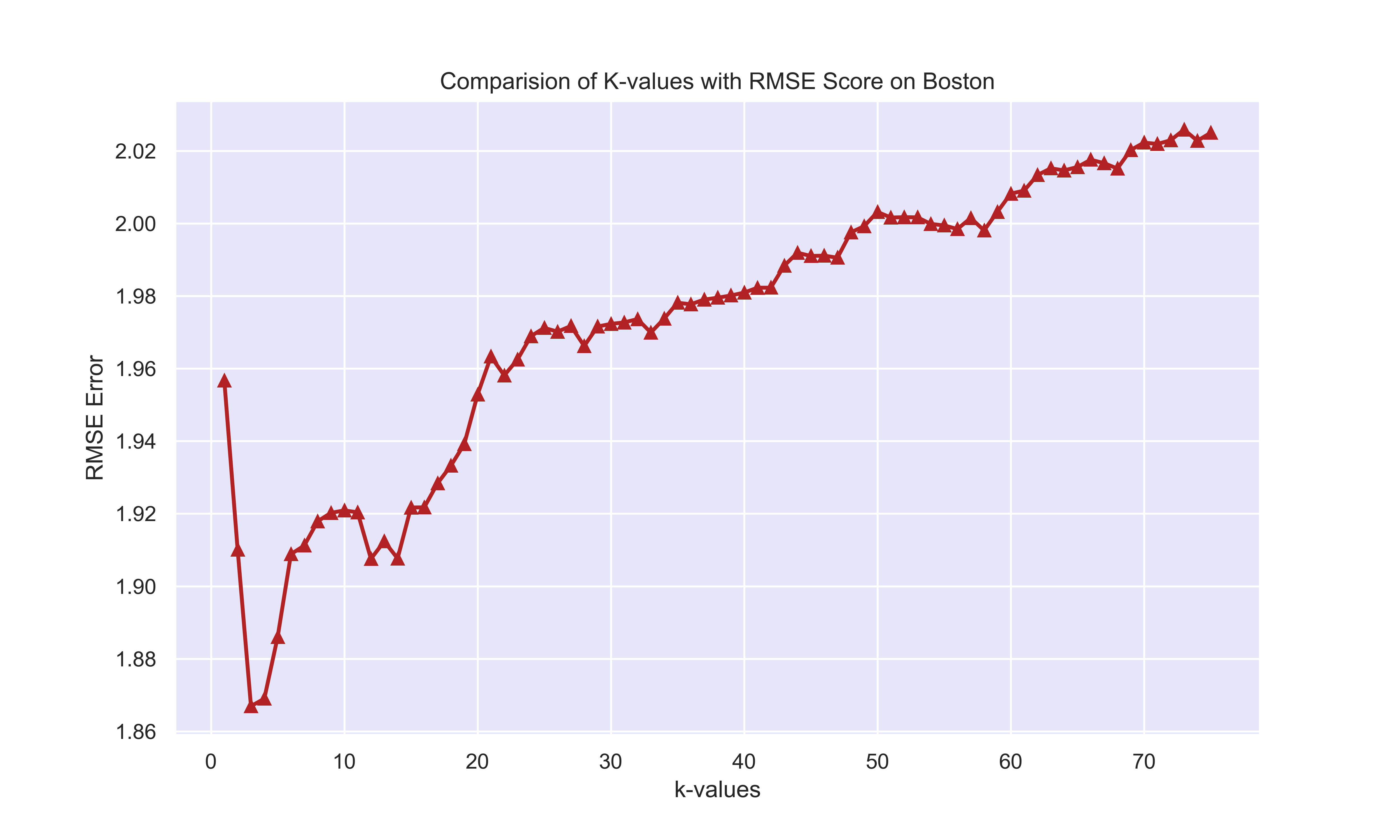}
    \caption{RMSE Score for Boston Housing Prices}
    \label{fig:1}
\end{figure}

It avoids considering the absolute value of the residuals and this attribute is effective in many mathematical calculations. In this metric, lower the value, better is the performance of the model. Standard Deviation is a measure of how spread out are the data points. Its equation has Variance square root and squared differences from the mean is variance average. So in order to get RMSE one uses Standard deviation formula with square root of average of squared residuals. The key point taken into consideration is, RMSE is most useful at the time of large errors. Absolute fit of the model on the data is done by RMSE. They are negatively-oriented scores which states, lower the values, better they are. After the prediction process the RMSE individually is considered for every single dataset over the set of varied k values. This can be visualized efficiently and it gives a detailed outlook towards the fluctuation of the k values. The Fig \ref{fig:1} is representation of the RMSE score calculated for Boston Housing Prices over the k values in range of 76. The graph gives the fluctuation in the values of the RMSE score where the lowest value of the RMSE is observed at a very early stage. Similarly the RMSE score for other 2 datasets can also be observed using visualizations.

\begin{figure}[htbp]
    \centering
    \includegraphics[scale=0.4]{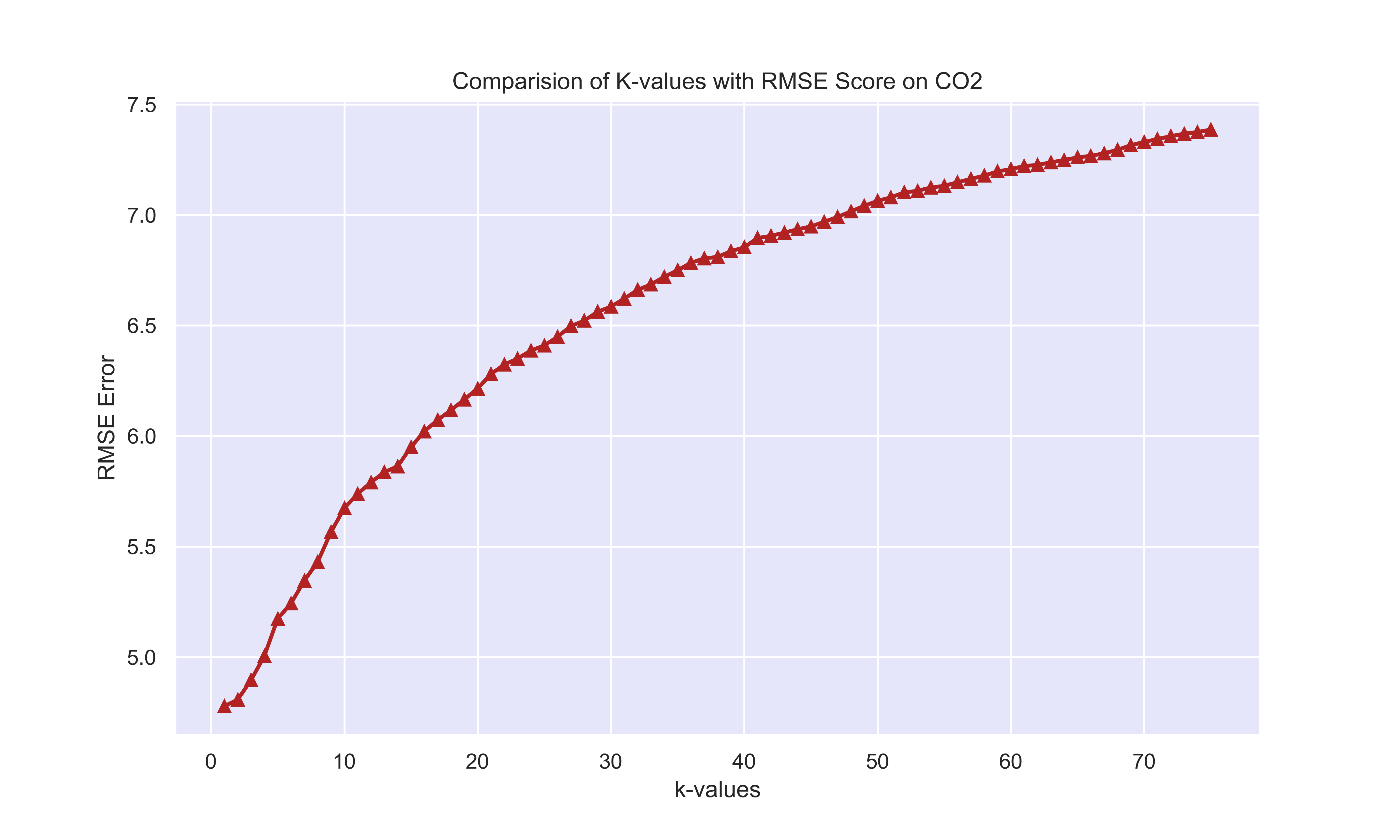}
    \caption{RMSE Score for Carbon Dioxide Emissions}
    \label{fig:2}
\end{figure}

The Fig \ref{fig:2} gives the representation of the RMSE values over the line in a quadratic curve style fashion. The lowest is observed before 5 values of k. This is an indication of transgression in the metric observation with respect to varied datasets.

\begin{figure}[htbp]
    \centering
    \includegraphics[scale=0.4]{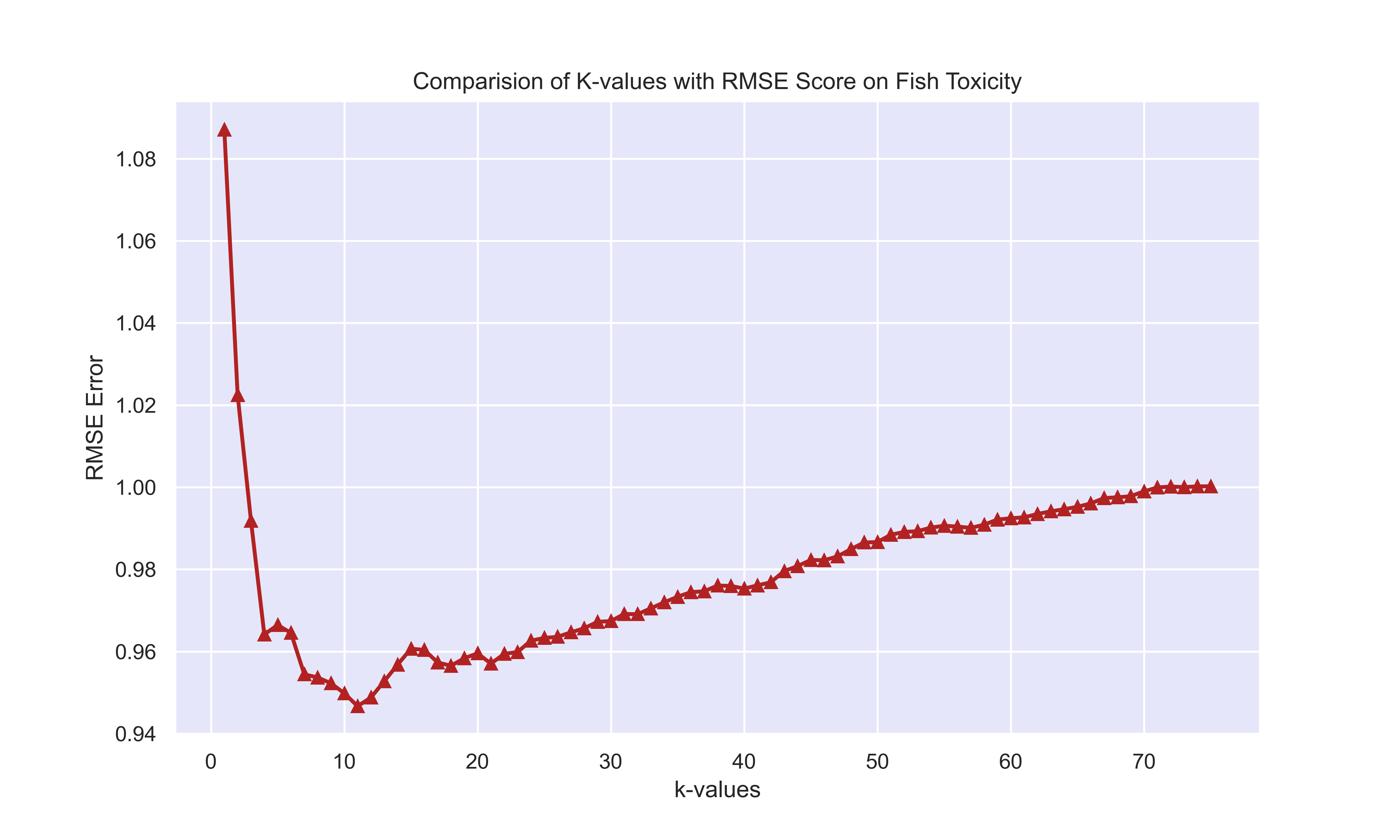}
    \caption{RMSE Score for LC50}
    \label{fig:3}
\end{figure}

The Fig \ref{fig:3} is the representation of the RMSE values over the 76 values of the k where the optimal score is observed around 10 and 20 values of the k. This is an indication of the score fluctuations and is not a robust metric to infer anything. 

\subsection{Goodness of Fit}
Goodness of Fit \citep{COLINCAMERON1997329} is the metric of accuracy for regression based models. The mathematical terminology for goodness of fit is known as $R^2$ or R-Squared Score. Coefficient of determination is the term given to it. R-Squared is a perfect indication of how effectively a models fits the given dataset. It can also be considered as an indication for relation of how effectively the regression line which are predicted values related to the actual test set of the data. The model gives a range between 0 and 1 as an indication of model performance given by metric. The values closer to 1 indicate the model is very good and vice versa. R-squared is a comparison of Sum of Squared Residuals (SSR) with Sum of Squared Totals (SST). SST is the calculation of summation performed over the perpendicular distance between the average line and its corresponding data points. SSR is the calculation of summation performed over the squares of perpendicular distance between best fit line and its data points. The equation for R-Squared is represented by the formula as

\begin{equation}
    R^2 = 1 - \frac{SSR}{SST}
\end{equation}

where the formula for SSR is given by

\begin{equation}
    SSR = \sum_i^n (\hat{y_i}- \overline{y_i})^2
\end{equation}

and the formula for SST is given by

\begin{equation}
    SST = \sum_i^n (y_i - \overline{y_i})^2
\end{equation}

The graphical representation of the varied values of R-Squared with respect to values of k ranging till 9 is given. The highest observation is found at the k value as 2 and it substantially becomes lower over the period of time which indicates the values of the data are less varied and broad distribution of the values of the k groups has more influence than getting into intricate details. The other observations can also be spot with other datasets.

\begin{figure}[htbp]
    \centering
    \includegraphics[scale=0.37]{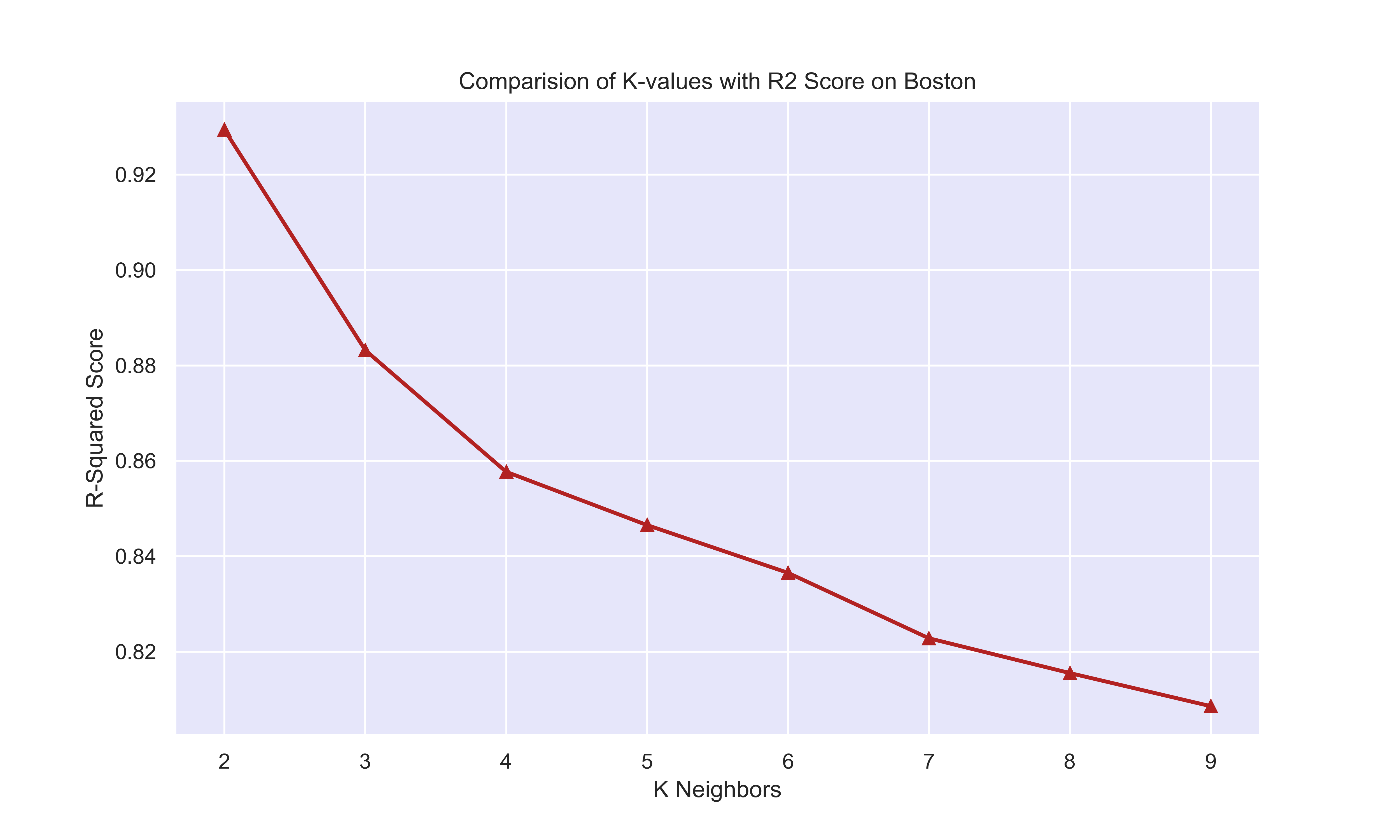}
    \caption{Goodness of Fit for Boston Housing Prices over k values}
    \label{fig:5}
\end{figure}

\begin{figure}[htbp]
    \centering
    \includegraphics[scale=0.37]{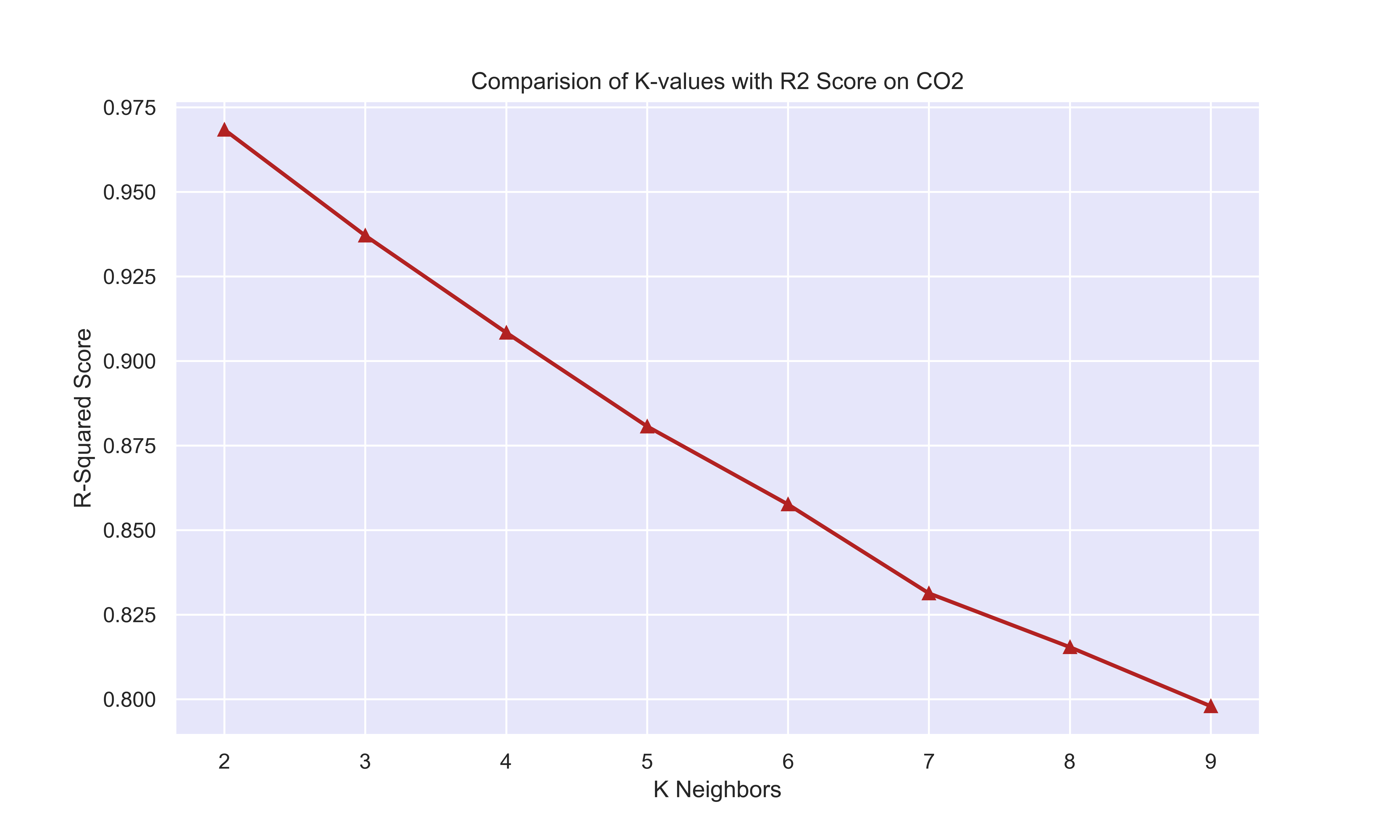}
    \caption{Goodness of Fit for CO2 Emissions over k values}
    \label{fig:6}
\end{figure}

The same type of observation in the graph can be seen with fig \ref{fig:6} and the influence of higher values of the k do not have any influence on the effect of the accuracy score. The less disparity between the classes have certain influence on the performance of the algorithm. This can be also checked on the final dataset for confirmation. 

The goodness of fit for the LC-50 has proved to be effective procedure to check the effect of the k values on the R-Squared accuracy metric. The higher values of k not necessarily indicate a better performance of the model. The optimal solution is important and the values of the k with certain effect make a lot of difference in most of the cases.

\begin{figure}[htbp]
    \centering
    \includegraphics[scale=0.37]{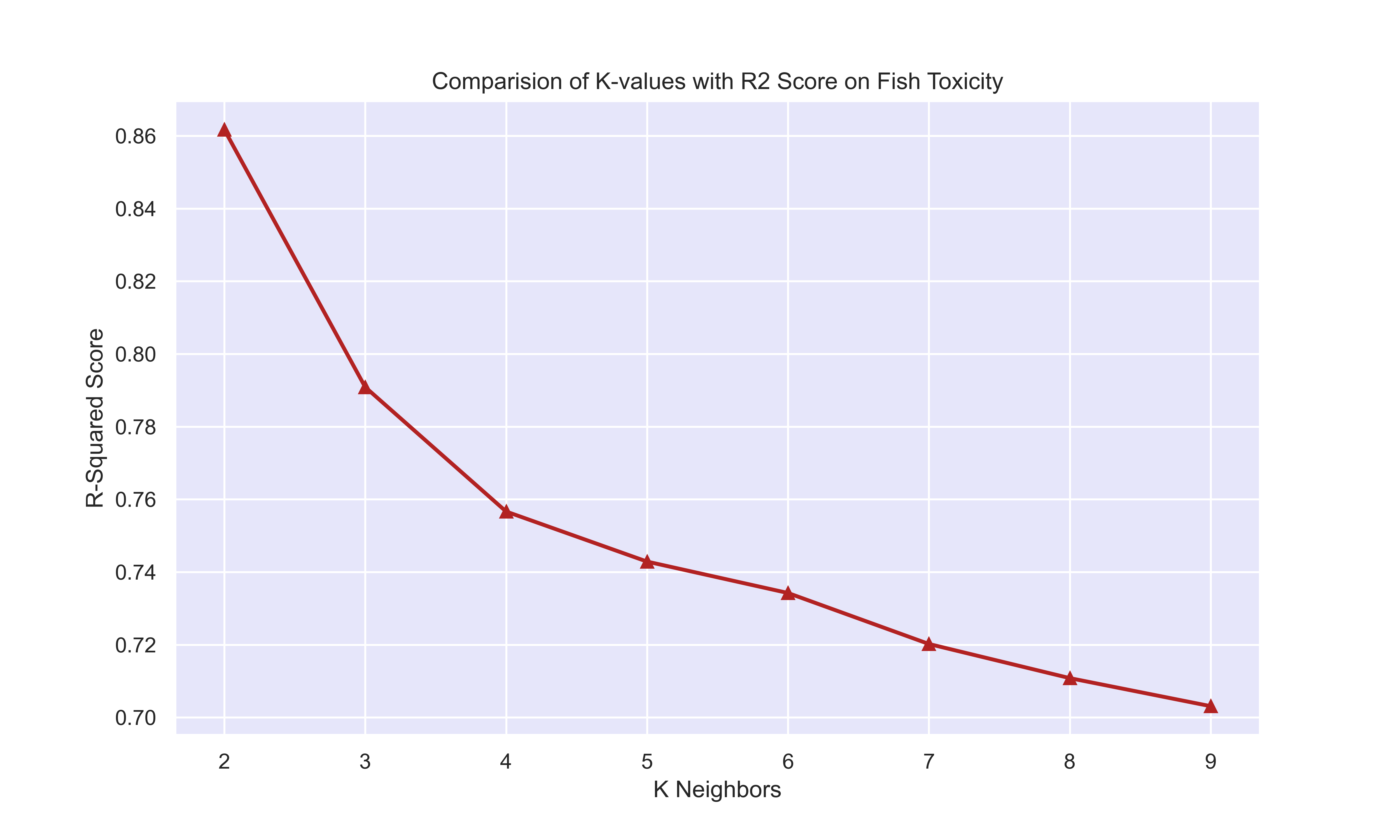}
    \caption{Goodness of Fit for LC-50 over k values}
    \label{fig:7}
\end{figure}

\section{Conclusion}
The main purpose of this paper emphasizes on the point of effect on metrics with respect to the fluctuations of k values for non-parametric regression based model. The non-parametric model we used for implementation is K Nearest Neighbor Regressor which is supervised learning model. The metrics we used to prove the subtle point were Root Mean Squared Error and R-Squared Goodness of Fit. The RMSE did not prove to be a better fit metric for proving the point as there were fluctuations in the values with respect to different datasets. The R-Squared on the other hand was able to prove the necessary point and performed very efficiently. It gives the optimal value of k in every single situation. The point was proved that the higher values of k do not influence the performance of the model and less distinctions between the values for separation holds more value than arbitrarily increasing the amount of k values. This is definitely not the end of the paper for focusing on such subtle observations and for sure opens many doors for new research which we will be glad to be a part of with our best belief and knowledge.

\bibliographystyle{unsrtnat}  
\bibliography{references}

\end{document}